\documentclass[sigconf,nonacm]{acmart}

\makeatletter
\renewcommand{\@ACM@checkaffil}{}
\makeatother

\AtBeginDocument{%
  }

\setcopyright{none}
\usepackage{ulem}
\usepackage{makecell}
\usepackage{multirow}
\usepackage{subfigure}
\usepackage{mathtools}
\usepackage{algorithm}
\usepackage{algpseudocode}
\usepackage{caption}
\begin{document}

\title{Reconstructing Fraudulent Strategies in Online Payments via Interpretable Inverse Reinforcement Learning}

\author{Huilin He$^{1,2,\dagger}$, Weiyu Han$^{2}$, Linbo Shao$^{1}$, Ruobing Zuo$^{3,2,\dagger}$, \\Xiao Tan$^{2,*}$, Shuai Chen$^{2}$, and Dawei Cheng$^{1,*}$} 
\affiliation{%
  \institution{
  $^1$ Tongji University, 
  $^2$ Ant International, 
  $^3$ East China Normal University
  }
}
\email{%
  {huilin3, 2634102, dcheng}@tongji.edu.cn, {hanweiyu.hwy, alex.tx, shuai.cs}@ant-intl.com, rbzuo@stu.ecnu.edu.cn
}

\renewcommand{\shortauthors}{He et al.}

\begin{abstract}
  Payment fraud causes financial losses and undermines users' trust in web payment services. Although fraud detection is a critical defense for payment security, complex and diverse fraudulent behaviors can still evade detection. Generating fraud samples for data augmentation can broaden detectors' coverage of different fraud patterns. However, existing methods matching the distribution of transaction attributes alone do not ensure that generated sequences preserve the sequential relationships among transaction choices, payment feedback, and subsequent behavioral adjustments. To this end, we propose PaySeqGen, an interpretable inverse reinforcement learning framework that reconstructs fraud behavior policies from fraudsters' observed payment trajectories. We design a mechanism for constructing candidate actions to support plausible payment sequence generation while enabling exploration. The framework also learns interpretable behavioral rewards that capture fraudsters' preferences across different transaction contexts. By alternating between reward learning and policy optimization, the framework reconstructs policies that generate new fraud sequences for detector augmentation and learns interpretable rewards supporting sequential fraud-pattern analysis in real trajectories. Experiments in real-world payment scenarios at Ant International, a leading global digital payment provider, show that PaySeqGen outperforms state-of-the-art generative augmentation methods. Reward-based trajectory analysis also identifies fraud patterns of practical value, helping analyst locating suspicious segments and examine multistep fraud tactics. Evaluation in the deployed system of online payment further confirms its effectiveness. 
\end{abstract}


\maketitle
\hypersetup{
  pdfauthor={Huilin He, Weiyu Han, Linbo Shao, Ruobing Zuo, Xiao Tan, Shuai Chen, Dawei Cheng}
}
{
  \renewcommand{\thefootnote}{} 
  \footnotetext{$*$ Corresponding Author.}
  \footnotetext{$\dagger$ Work done during internships at Ant International.}
  \renewcommand{\thefootnote}{\arabic{footnote}} 
}

\section{Introduction}

\begin{figure}
  \centering
  \includegraphics[width=\linewidth]{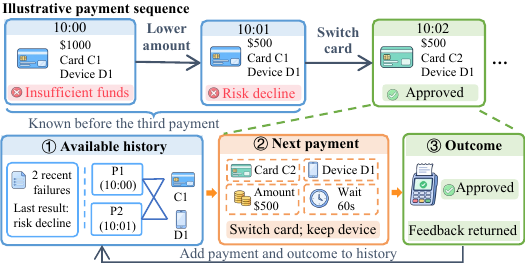}
  \vspace{-15pt}
  \caption{An illustrative online fraud payment sequence and its decision context shaped by feedback.}
  \label{fig:pay seq modeling}
 \vspace{-15pt}
\end{figure}

Online payment systems underpin e-commerce and digital services, making their security a critical concern for users and merchants~\cite{alfalahi2024bankruptcy}. Payment fraud causes financial losses for users, merchants, and payment providers, raises risk management and dispute resolution costs, damages platform reputations, and undermines trust in online payment services~\cite{mate2019frauds, consumer2024sentinel}. Fraud detection therefore serves as a key line of defense~\cite{cheng2020spatio}. However, fraudulent behavior takes diverse forms across different payment contexts, and limited training data may not adequately capture complex fraud patterns and their variations. This limited coverage constrains detectors' ability to learn and reliably recognize such patterns in practice~\cite{qiao2025online,d2025synthetic}.

Generating additional payment sequences from observed fraudulent behavior provides a way to expand this coverage by introducing behavioral variations into detector training. Existing sequence generation methods typically model transaction attribute distributions and temporal dependencies~\cite{NEURIPS2019_c9efe5f2,yuan2024diffusion,10.1007/978-3-031-26422-1_8,zhang2024mixedtype}. However, matching observed sequences alone does not necessarily ensure that generated decisions remain coherent as the history evolves and new feedback becomes available. In web payment settings, each payment attempt updates the transaction history, and its outcome can influence subsequent decisions about payment amounts, intervals, and card and device use, as illustrated in Figure~\ref{fig:pay seq modeling}. Fraud sequence generation therefore calls for explicit modeling of this interaction process: selecting the next payment action based on the current history, incorporating the resulting payment feedback into that history, and making subsequent decisions that reflect adaptive fraud strategies while respecting payment constraints.

Such feedback-driven sequential decision-making is naturally modeled using reinforcement learning (RL), which learns policies by maximizing expected cumulative reward and has also been applied to sequence generation~\cite{3298483.3298649,11434559}. In our setting, however, the reward is unknown. Handcrafting rewards requires specifying in advance which payment behaviors should be encouraged. These predefined preferences may not align with those reflected in real demonstrations and can bias the learned policy. Inverse reinforcement learning (IRL)~\cite{1620270.1620297,pmlr-v48-finn16,garg2021iqlearn} offers an alternative by inferring rewards that explain demonstrated behavior. In our setting, observed payment trajectories can serve as demonstrations for learning a behavioral reward and reconstructing a policy that generates new payment sequences. A decomposable reward representation further exposes the scoring contributions of individual behavioral factors, providing interpretable evidence for analyzing sequential fraud patterns. Prior work on rumor detection has used IRL to reconstruct graph-structure attack  policies~\cite{lyu2024enhancing}, providing a methodological precedent for this approach. However, using interpretable IRL to reconstruct fraudulent strategies in online payments remains underexplored.

Applying this approach to web payments presents two challenges. First, candidate action construction must balance behavioral coherence and exploration. Decisions to reuse or switch cards and devices must account for transaction history and payment feedback, as must decisions about payment amounts, intervals, and other attributes. Generating these fields independently can disrupt behavioral coherence~\cite{11434559}. Meanwhile, the combinatorial space of payment actions is too large to enumerate. The challenge is therefore to cover historically supported behavioral combinations with a finite candidate set while allowing exploration beyond combinations observed in demonstrations. Second, rewards must capture context-dependent behavioral preferences~\cite{pmlr-v267-ke25b}. The same action, such as switching cards or increasing a payment amount, may reflect different behavioral patterns depending on preceding transactions and payment outcomes. Scoring action attributes in isolation makes it difficult to distinguish context-dependent preferences.

To address these challenges, we propose PaySeqGen, an interpretable inverse reinforcement learning framework for reconstructing payment fraud policies from observed trajectories. At each step, the state summarizes payment history and feedback, and the action specifies the next payment. To balance behavioral coherence and exploration within a finite action set, we combine history-guided sampling with exploratory sampling. History-guided sampling draws joint changes from historical transitions matching contexts and feedback, adapting them to current history. Exploratory sampling recombines payment components under resource and time constraints. To capture context-dependent behavioral preferences, we learn a linear behavioral reward over base features and feedback--action interactions, allowing action scores to vary with preceding payment feedback. We alternate reward learning with policy optimization to reconstruct payment fraud strategies, using a fixed payment feedback interface adapted from the platform's risk-control system to provide outcomes for generated payments. After training, the reconstructed policy generates fraud payment sequences to augment detector training. The learned interpretable reward helps characterize sequential fraud patterns in observed trajectories.

Our contributions are summarized as follows:

\begin{itemize}
    \item We propose PaySeqGen, an interpretable IRL framework for reconstructing fraudulent payment strategies from observed trajectories, supporting both sequence generation for detector augmentation and reward-guided analysis of sequential fraud patterns.
    \item We design a candidate construction mechanism that combines history-guided and exploratory sampling to balance behavioral coherence and exploration, and a decomposable linear reward with feedback--action interactions to capture context-dependent behavioral preferences.
    \item We evaluate PaySeqGen on large-scale data from a real-world online payment system operated by Ant International\footnote{https://www.ant-intl.com/en/}, a leading global digital payment and financial technology provider. Offline evaluations and online A/B tests demonstrate PaySeqGen's augmentation benefits. Reward-guided analysis also provides traceable evidence for fraud investigation. These results demonstrate PaySeqGen's practical value in strengthening web payment security.
\end{itemize}

\section{Preliminaries}

\subsection{Reinforcement Learning and Actor-Critic}

In reinforcement learning, a policy $\pi_\omega(a_t\mid s_t)$
selects actions at each step and interacts with environment dynamics $P$ to
generate a trajectory $\tau=(s_1,a_1,\ldots,s_T,a_T)$.
Policy learning maximizes the expected cumulative reward:
\begin{equation}
\label{eq:rl_objective}
J(\omega)
=
\mathbb E_{\tau\sim\pi_\omega,P}
\left[
\sum_{t=1}^{T}r(s_t,a_t)
\right].
\end{equation}
Here, $\omega$ denotes the policy parameters, $P$ is the environment transition distribution, $\tau$ is a trajectory containing $T$ decision steps, and $r$ is the reward function. 
Policy updates based on sampled returns can have high variance~\cite{sutton1999policy}. We adopt Actor--Critic~\cite{konda1999actor} to provide a learned value baseline for these updates. The Actor $\pi_\omega$ outputs action probabilities, while the Critic $V_\psi(s_t)$ estimates the expected return from $s_t$ under the current policy, with $\psi$ denoting the Critic parameters. Let $\widehat G_t$ denote a return estimate based on the selected action and subsequent interactions. The estimated advantage is defined as:
\begin{equation}
\label{eq:advantage}
\widehat A_t
=
\widehat G_t - V_\psi(s_t).
\end{equation}
The advantage measures how much better the selected action performs relative to the current policy's average behavior at the same state. The Actor uses this signal to favor actions with better-than-expected returns, while the Critic is trained to predict the return targets $\widehat G_t$ computed from sampled trajectories.

\subsection{Maximum Entropy IRL}

Inverse reinforcement learning (IRL) assumes that expert behavior is optimal or near-optimal under an unknown reward function and aims to recover a reward that explains the observed behavior. 
To account for variability in expert behavior, maximum entropy IRL models a stochastic policy that favors higher cumulative rewards while retaining behavioral diversity~\cite{1620270.1620297}.
Let $r_\theta(s,a)$ denote a parameterized reward function and $\pi_\theta$ the stochastic policy induced by this reward under the maximum-entropy model.
Given an expert demonstration set $\mathcal D$, reward learning can be formulated as maximizing the log-likelihood of the demonstrated actions:
\begin{equation}
\label{eq:irl_likelihood}
\theta^*
=
\arg\max_\theta
\sum_{\tau\in\mathcal D}\sum_t
\log \pi_\theta(a_t^E \mid s_t^E),
\end{equation}
where $(s_t^E,a_t^E)$ is a demonstrated state--action pair in trajectory $\tau$. This objective adjusts the reward parameters so that the induced policy assigns higher probabilities to expert actions, thereby encoding demonstrated behavioral preferences as explicit reward scores. Our framework follows this perspective through sampled feature-expectation matching and alternating reward and policy updates.

\begin{figure*}[t!]
  \centering
  \includegraphics[width=\textwidth]{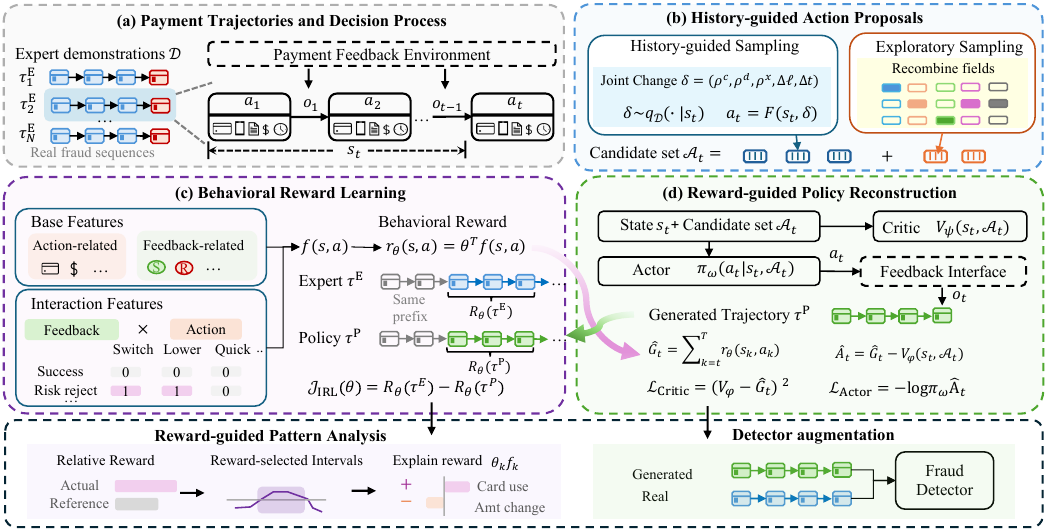}
  \vspace{-15pt}
  \caption{Overview of PaySeqGen. (a) Model payment trajectories as a sequential decision process. (b) Construct candidate actions through history-guided sampling and exploratory sampling. (c) Learn behavioral rewards with feedback--action interactions, supporting fraud-pattern analysis and guiding fraud policy reconstruction. (d) Optimize the policy using learned rewards to generate payment sequences for detector augmentation.}
  \label{fig:overview}
\end{figure*}

\section{Methodology}
\subsection{Overview}
As shown in  Figure~\ref{fig:overview}, PaySeqGen learns an
interpretable behavioral reward and reconstructs a fraud payment policy from real demonstrations. The policy selects the next payment based on the current history, while the reward evaluates whether this choice reflects the behavioral preferences in the demonstrations. To support these roles, candidate construction balances behavioral coherence and exploration, while behavioral reward learning captures how action preferences vary with prior feedback. Within this process, reward learning and policy reconstruction alternate: differences between demonstrated and generated behavior inform reward updates, which in turn guide policy optimization. After training, the reconstructed policy generates sequences for detector augmentation, while the learned reward supports fraud-pattern analysis.

\subsection{Payment Trajectories and Decision Process}

Let a real trajectory be $\tau=\{(a_t,o_t)\}_{t=1}^{T}$, where $a_t$ is a payment action and $o_t$ is the feedback returned after execution. 
We represent the next payment event choice as an action:
\begin{equation}
\label{eq:payment_action}
a_t=(c_t,d_t,x_t,m_t,\Delta t_t)
\end{equation}
where $c_t$ and $d_t$ denote the card and device, respectively; $x_t$ contains card attributes and payment-context attributes, such as country and terminal information; $m_t$ is the payment amount; and $\Delta t_t$ is the waiting time since the previous payment.
The state $s_t$ is represented as the ordered history of payments and their observed feedback:
\begin{equation}
\label{eq:decision_state}
s_t=\bigl((a_1,o_1),\ldots,(a_{t-1},o_{t-1})\bigr).
\end{equation}
Real payment transactions receive feedback from the payment system, whereas generated actions require simulated feedback, since underlying business states, such as account balances, are not fully observable. We therefore use a payment feedback interface adapted from the platform’s risk-control system with a multiclass output head. This interface approximates the distribution of payment outcomes, including success, rejection, and other failure categories, and remains fixed during reward and policy learning:
\begin{equation}
\label{eq:feedback_transition}
\begin{aligned}
o_t &\sim P_{\mathrm{env}}(\cdot\mid s_t,a_t), \\
s_{t+1} &= \operatorname{Update}(s_t,a_t,o_t).
\end{aligned}
\end{equation}
Here, $P_{\mathrm{env}}$ is the conditional distribution of payment feedback, and $\operatorname{Update}$ appends the current payment and its feedback to the history. Feedback for the current payment becomes available only after execution and is used in the next decision.

\subsection{History-guided Action Proposals}

The action candidate set $\mathcal A_t$ defines the payment choices available to the policy at step $t$, shaping both sequence coherence and the scope of exploration. Constructing this set is challenging because payment attributes form a large combinatorial space, while independent field sampling can disrupt their dependencies. We therefore combine joint candidates guided by real payment histories with constrained random exploration, preserving observed attribute dependencies while expanding the behavioral combinations available to the policy.
We first construct history-guided candidates from joint changes observed between consecutive payments in the demonstration set $\mathcal D$. Each change is represented as:
\begin{equation}
\label{eq:joint_transition}
\delta_i
=
(\rho_i^c,\rho_i^d,\rho_i^x,\Delta\ell_i,\Delta t_i),
\end{equation}
where $\rho_i^c$ and $\rho_i^d$ describe card and device usage, such as retaining the current resource, returning to a previously used resource, or introducing a new one. $\rho_i^x$ records whether the payment context is retained or replaced, together with the corresponding attribute template. $\Delta\ell_i$ is the change in log amount $\log(1+m)$, where $m$ is the payment amount, and $\Delta t_i$ is the interval between payments. These components are extracted from the same pair of consecutive payments and sampled jointly to preserve co-occurring behavioral changes.
At state $s_t$, we sample $K_h$ joint changes by matching the latest payment feedback, amount bin, and payment context, and adapt them to the current history:
\begin{equation}
\label{eq:historical_action_sampling}
\begin{aligned}
\delta_{t,j}
&\sim q_{\mathcal D}(\cdot\mid s_t), \\
a_{t,j}^{\mathrm{hist}}
&=F(s_t,\delta_{t,j}),
\qquad j=1,\ldots,K_h.
\end{aligned}
\end{equation}
Here, $q_{\mathcal D}$ samples joint changes from demonstrations under matching payment conditions. The mapping $F$ adapts the sampled changes to the current resources, attributes, and time budget. The resulting candidates form $\mathcal A_t^{\mathrm{hist}}$.
For exploration, we draw $K_e$ proposals
$\tilde a_{t,j}\sim q_{\mathrm{exp}}(\cdot\mid s_t)$
and construct candidates as:
\begin{equation}
\label{eq:exploration_sampling}
\begin{aligned}
q_{\mathrm{exp}}(\tilde a\mid s_t)
&=
\prod_{u\in\{c,d,x,m,\Delta\}}
q_u(\tilde a_u\mid s_t), \\
a_{t,j}^{\mathrm{exp}}
&=F_{\mathrm{exp}}(s_t,\tilde a_{t,j}),
\qquad j=1,\ldots,K_e.
\end{aligned}
\end{equation}
Here, $q_u$ samples a payment component without historical transition matching, and $F_{\mathrm{exp}}$ enforces resource-consistency and time-budget constraints. These candidates form $\mathcal A_t^{\mathrm{exp}}$. The final candidate list is
$\mathcal A_t=\operatorname{Concat}(\mathcal A_t^{\mathrm{hist}},\mathcal A_t^{\mathrm{exp}})$.

\subsection{Behavioral Reward Learning}

\subsubsection{Decomposable Behavioral Rewards}

To make learned behavioral preferences interpretable, we define a linear reward over a set of fixed behavioral features:
\begin{equation}
\vspace{-3pt}
\label{eq:contextual_linear_reward}
\begin{aligned}
f(s,a)
&=[b(s,a);\chi(s,a)], \\
r_\theta(s,a)
&=\theta^\top f(s,a).
\end{aligned}
\vspace{-3pt}
\end{equation}
Here, $b(s,a)$ summarizes amounts, payment intervals, resource usage, and historical feedback. Preferences for the same action may vary with prior payment feedback. We therefore construct the interaction vector $\chi(s,a)$ from products of feedback-related and action-related features.
For example, 
\begin{equation}
\label{eq:rejection_card_switch}
\begin{aligned}
\chi_{\mathrm{reject,switch}}(s_t,a_t)
&=
\mathbf 1[o_{t-1}=\mathrm{risk\ rejection}] \\
&\quad\cdot
\mathbf 1[c_t\ne c_{t-1},\,d_t=d_{t-1}]
\end{aligned}
\end{equation}
captures switching cards while retaining the device after a risk rejection.
The feature definitions are fixed and shared by the policy and reward. The policy uses them for nonlinear scoring, whereas the reward learns their linear weights. Each contribution $\theta_k f_k(s,a)$ can therefore be traced to a specific behavioral factor, supporting subsequent fraud-pattern analysis.

\subsubsection{Reward Optimization}

We use an observed payment prefix as a shared starting history to compare expert and policy behavior under the same initial conditions. The expert branch replays the recorded continuation, while the policy branch generates an alternative. The gap between their behavioral feature expectations provides the signal for updating the reward weights.
For the $L=T-p$ decisions following a prefix of $p$ payments, let
$\bar f(\tau)=L^{-1}\sum_{t=p+1}^{T}f(s_t,a_t)$
denote the average feature vector over the continuation. The expert and policy feature expectations are:
\begin{equation}
\vspace{-2.5pt}
\begin{aligned}
\label{eq:feature_expectations}
\mu^E&=\mathbb E_{\tau\sim\mathcal D}[\bar f(\tau)]\\
\mu^\pi&=\mathbb E_{\tau\sim\pi_\omega,P_{\mathrm{env}}}[\bar f(\tau)],
\end{aligned}
\vspace{-2.5pt}
\end{equation}
where $\mathcal D$ is the demonstration set. Following feature-expectation matching in
IRL~\cite{abbeel2004apprenticeship,1620270.1620297}, we hold the sampled trajectories fixed and update the reward by minimizing:
\begin{equation}
\label{eq:irl_loss}
\mathcal L_{\mathrm{IRL}}(\theta)
=
\theta^\top(\mu^\pi-\mu^E)
+\frac{\lambda}{2}\|\theta\|_2^2.
\end{equation}
We optimize the reward weights by gradient descent, with each weight clipped to $[-w_{\max},w_{\max}]$.
The regularization term discourages large weights, while clipping imposes an explicit bound on each learned reward coefficient. The feature-matching term increases weights for features underrepresented in generated trajectories and decreases those overrepresented relative to the expert demonstrations. The updated reward then guides the next policy optimization step. Through alternating reward and policy updates, the learned weights identify which behavioral factors and feedback--action combinations to encourage or discourage to reproduce demonstrated feature statistics, expressing context-dependent preferences as explicit, decomposable scores.

\subsection{Reward-guided Policy Reconstruction}

\subsubsection{Policy representation and initialization.}
Given the candidate list $\mathcal A_t$, the policy selects
the next payment from its $K$ candidates. For each candidate
$a_{t,j}$, we combine the behavioral features defined above
with learned attribute encodings into a vector:
$\mathbf h_{t,j}
=[f(s_t,a_{t,j});e_a(a_{t,j});e_s(s_t)]$.
Here, semicolons denote concatenation, $e_a$ encodes candidate
attributes, and $e_s$ mean-pools learned attribute embeddings
over the payment history.
The Actor scores all candidates with the same multilayer
perceptron (MLP) and normalizes their scores into selection
probabilities:
\begin{equation}
\label{eq:actor_policy}
\pi_\omega(j\mid s_t,\mathcal A_t)
=
\frac{
\exp\!\left(\operatorname{MLP}_{\omega}(\mathbf h_{t,j})\right)
}{
\sum_{k=1}^{K}
\exp\!\left(\operatorname{MLP}_{\omega}(\mathbf h_{t,k})\right)
}.
\end{equation}
The Critic estimates the expected remaining return under the
current policy as
$V_\psi(s_t,\mathcal A_t)
=\operatorname{MLP}_{\psi}
\left(K^{-1}\sum_{j=1}^{K}\mathbf h_{t,j}\right)$.
Conditioning the Critic on the candidate list allows its baseline
to account for the payment choices available at each step.
The Actor and Critic share the attribute encoders and use
separate MLP heads. Their parameter sets, $\omega$ and $\psi$,
include the shared encoder parameters.
Before autonomous generation, we initialize the Actor through behavior cloning: the demonstrated next payment is included in each candidate set, and the policy maximizes its log-probability under the corresponding real history. This provides an initial payment-selection policy for subsequent reinforcement learning.

\subsubsection{Reward-guided policy optimization.}
With the current behavioral reward $r_\theta$ fixed, the policy samples the next payment from $\mathcal A_t$. The payment feedback interface provides an outcome, which is added to the history together with the selected action. We add an entropy bonus to encourage exploration among candidates and normalize by the number of generated steps:
\begin{equation}
\label{eq:training_reward}
\tilde r_t
=
\frac{
r_\theta(s_t,a_t)
-\alpha\log\pi_\omega(j_t\mid s_t,\mathcal A_t)
}{L},
\end{equation}
where $L$ is the number of generated decision steps following the initial prefix, and $\alpha$ controls the strength of entropy regularization. We use generalized advantage estimation (GAE)~\cite{schulman2015high} to estimate $\widehat A_t$ by combining sampled rewards with Critic predictions over multiple horizons. This balances variability in sampled returns against value-estimation bias, helping stabilize policy updates. The corresponding return targets $\widehat G_t$ are computed using Eq.~\eqref{eq:advantage}. We then minimize the joint loss:
\begin{equation}
\label{eq:rl_training_loss}
\begin{aligned}
\mathcal L_{\mathrm{RL}}(\omega,\psi)
&=
-\mathbb E_t
\left[
\log\pi_\omega(j_t\mid s_t,\mathcal A_t)
\,\widehat A_t
\right] \\
&\quad+
\beta_V\mathbb E_t
\left[
\left(
V_\psi(s_t,\mathcal A_t)-\widehat G_t
\right)^2
\right],
\end{aligned}
\end{equation}
where $\mathbb E_t$ averages over generated decision steps
in the sampled batch, and $\beta_V$ weights the Critic loss. The advantages and return targets are held fixed during each update. The first term increases the probability of choices with better-than-expected multistep returns, while the second improves the Critic's return estimates.

\subsection{Alternating Training and Applications}

We initialize the policy by behavior cloning on demonstrations and set the reward weights to zero. Training then alternates between two steps: updating the reward using feature-expectation differences between expert and current policy trajectories, and fixing the reward while sampling fresh trajectories for policy optimization. Each policy update changes the generated distribution, which in turn informs the next reward update.

\subsubsection{Fraud sequence generation and detector augmentation.}
After training, we use the reconstructed fraud policy to generate synthetic fraudulent payment sequences for detector augmentation. Each sequence begins with an initial payment assembled from training-data components and uses fresh user and resource identifiers to avoid adding synthetic payments to real users' existing histories. The policy then generates subsequent payments using feedback from the payment feedback interface. The final payment serves as the fraud-labeled target, with preceding payments providing its historical context. The synthetic sequences are combined with real training samples for detector augmentation.

\subsubsection{Reward-guided Fraud Pattern Analysis.}
With the learned reward fixed, we replay real trajectories to identify informative behavioral intervals for fraud pattern analysis. At each historical state $s_t$, we reuse the candidate construction mechanism from policy generation to obtain $\mathcal A_t$. We define the relative reward $\Delta r_t$ as the difference between the observed action's reward and the mean reward of the candidate actions:
\begin{equation}
\label{eq:relative_reward}
\begin{aligned}
\Delta r_t
&= r_\theta(s_t,a_t^E)-
\frac{1}{|\mathcal A_t|}
\sum_{a\in\mathcal A_t}
r_\theta(s_t,a).
\end{aligned}
\end{equation}
Here, $a_t^E$ is the observed payment action. A larger $\Delta r_t$ indicates a stronger learned preference for this choice relative to the candidate alternatives under the same history.
Since the reward is learned from fraud-related demonstrations, predominantly positive scores across consecutive payments provide evidence of sustained alignment with the modeled behavioral preferences. We therefore select such intervals as candidate segments for fraud-pattern analysis. Signed feature contributions identify which factors raise or lower the observed actions' rewards relative to the candidate mean. Together with recorded amounts, resource changes, and payment feedback, these contributions help characterize the sequential behavior and contextual preferences within each selected interval.

\section{Experiments}

\subsection{Experiment Setup}

\subsubsection{Datasets.}

We conduct experiments on real-world data collected from a large online payment platform operated by Ant International. Our dataset covers approximately 1.29 million users and 26.28 million distinct payment events across 3.46 million labeled detection examples. Each example corresponds to a user's payment sequence ending in a target payment, and its binary label indicates whether that target payment is fraudulent. Target payments span March 2025 to January 2026. The records include payment amounts and timing, card and device usage, and available payment feedback. All detectors share the same chronological training, validation, and test splits. Expert demonstrations are drawn from training examples with fraudulent targets; action-proposal templates and sampling statistics are also derived exclusively from the training split.



\subsubsection{Target Models}

We evaluate the utility of generated samples for augmenting four fraud detectors with different architectures. Our primary detector, referred to as RT, is a Transformer-based~\cite{vaswani2017attention} payment fraud detection model deployed in our production risk control system.
Given the sequential nature of our task, we also include TabFormer~\cite{padhi2021tabular}, which models payment sequences using a hierarchical Transformer architecture at the field and payment-event levels, with masked-attribute pretraining followed by fine-tuning for fraud detection. We also evaluate GraphSAGE~\cite{hamilton2017inductive}, motivated by the widespread use of graph neural networks in fraud detection~\cite{cheng2025graph,liu2026empowering}. This allows us to assess whether the benefits of generated samples extend to detectors that model relationships among payments. Finally, we include XGBoost~\cite{chen2016xgboost}, a well-established baseline in production fraud detection systems, to assess whether the benefits of generated samples extend to tree-based fraud detectors.

\subsubsection{Generation Baselines}

We compare PaySeqGen with two categories of generation baselines.
Distribution-based methods include CTGAN~\cite{xu2019modeling} for conditional adversarial generation of mixed-type attributes, Diffusion-TS~\cite{yuan2024diffusion} for diffusion-based sequence generation with temporal decomposition, and TabDiff~\cite{shi2025tabdiff} for joint diffusion over continuous and discrete fields.
Policy-based methods include Behavior Cloning (BC)~\cite{osa2018algorithmic} and MoE-BiEntIRL~\cite{lyu2024enhancing}.  BC learns a payment policy through supervised imitation of expert actions. It shares PaySeqGen's policy architecture and candidate construction mechanism but is trained independently by maximizing demonstrated-action likelihood, without IRL or RL updates. MoE-BiEntIRL combines IRL with a mixture-of-experts formulation.
We additionally include a \textit{Real-fraud replay} control, which adds an equal number of resampled real fraud examples. This control helps distinguish the benefits of generating behavioral variations from the effect of simply increasing the weight of fraud examples during detector training. All generation methods use the same training data sources and generation lengths.

\begin{table*}[t!]
\setlength{\tabcolsep}{2pt}
\centering
\caption{Performance changes across four fraud detectors under different augmentation methods. Values are percentage-point differences from training without augmentation. Boldface and underlining indicate the best and second-best results per column.
}
\vspace{-5pt}
\label{tab:augmentation_comparison}

\begin{tabular}{@{}cl|ccc|ccc|ccc|ccc@{}}
\toprule
\multicolumn{2}{c|}{Detectors}
& \multicolumn{3}{c|}{RT}
& \multicolumn{3}{c|}{TabFormer}
& \multicolumn{3}{c|}{GraphSAGE}
& \multicolumn{3}{c}{XGBoost} \\
\midrule
\multicolumn{2}{c|}{Metrics}
& $\Delta$AUC & $\Delta$Recall@2 & $\Delta$KS
& $\Delta$AUC & $\Delta$Recall@2 & $\Delta$KS
& $\Delta$AUC & $\Delta$Recall@2 & $\Delta$KS
& $\Delta$AUC & $\Delta$Recall@2 & $\Delta$KS \\
\midrule

Control & Fraud replay
& -0.046 & +1.094 & \uline{+0.166}
& -0.092 & \uline{-1.679} & -0.042
& +0.036 & +1.098 & +0.416
& \textbf{+0.109} & \textbf{+0.325} & \textbf{+0.074} \\

\midrule

\multirow{5}{*}{Baselines}
& BC
& -0.141 & -0.360 & +0.055
& +0.089 & -2.972 & -0.097
& \uline{+0.681} & \uline{+3.375} & +0.975
& -0.195 & -1.266 & -0.354 \\

& CTGAN
& -0.036 & \uline{+1.559} & -0.169
& +0.148 & -2.218 & +0.170
& +0.066 & +1.370 & +0.934
& -0.025 & -0.193 & -0.304 \\

& Diffusion-TS
& -0.189 & +0.871 & -0.610
& -0.042 & -1.923 & -0.423
& +0.343 & +1.454 & \uline{+1.056}
& -0.095 & -0.976 & -0.232 \\

& TabDiff
& \uline{+0.169} & +0.769 & +0.108
& \uline{+0.198} & -2.000 & \uline{+0.295}
& -0.251 & +0.111 & -0.517
& -0.093 & -0.103 & -0.628 \\

& MoE-BiEntIRL
& -0.051 & +1.036 & -0.042
& +0.149 & -2.274 & -0.130
& +0.360 & +1.951 & +0.697
& -0.030 & \uline{-0.004} & -0.428 \\

\midrule

Ours & FULL
& \textbf{+0.393} & \textbf{+1.621} & \textbf{+0.218}
& \textbf{+0.328} & \textbf{-1.613} & \textbf{+0.451}
& \textbf{+0.878} & \textbf{+4.561} & \textbf{+1.686}
& \uline{+0.062} & -0.047 & \uline{-0.032} \\

\midrule

\multirow{3}{*}{Ablations}
& w/o coh.
& +0.175 & +1.171 & +0.065
& +0.089 & -1.700 & +0.186
& +0.654 & +1.426 & +1.332
& +0.034 & -0.411 & -0.054 \\

& w/o expl.
& +0.272 & +0.568 & -0.071
& +0.100 & -1.872 & +0.006
& +0.596 & +1.165 & +1.042
& +0.042 & -0.704 & +0.092 \\

& w/o inter.
& -0.257 & +0.653 & -0.151
& +0.240 & -1.929 & +0.343
& +0.588 & +1.250 & +1.454
& -0.022 & -0.231 & -0.187 \\

\bottomrule
\end{tabular}
\end{table*}

\subsubsection{Evaluation Metrics}
We evaluate detector augmentation using the area under the ROC curve (AUC), Recall@2\%, and the Kolmogorov--Smirnov (KS) statistic. AUC measures the detector's ability to rank fraudulent payments above legitimate ones. Recall@2 measures the proportion of all fraudulent samples captured among the top 2\% of samples ranked by predicted risk. KS measures the maximum difference between the true positive rate and the false positive rate across all thresholds,  reflecting class separation. Higher values indicate better performance for all three metrics.

\subsubsection{Implementation Details}

Experiments are conducted on a server using CUDA~12.4 for acceleration and PyTorch~2.4.0~\cite{paszke2019pytorch} for training and evaluation. For each generator and random seed, we augment the detector training set with synthetic fraud samples equivalent to approximately 2.5\% of its original size. Original training examples and the validation and test sets remain unchanged.
PaySeqGen constructs 64 action candidates per step, comprising 48 history-guided candidates and 16 exploration candidates. We use a policy hidden dimension of 128 and a training batch size of 256. The policy and reward learning rates are $3\times10^{-4}$ and $0.03$, respectively. For generalized advantage estimation, we set the discount factor to $\gamma=1$ and the smoothing parameter to $\lambda_{\mathrm{GAE}}=0.95$. Baselines and detectors follow their original implementations or reported settings, with adaptations to payment sequences.

\subsection{Improvement of Detection Performance with Generated Samples}

Table~\ref{tab:augmentation_comparison} shows that PaySeqGen achieves the best or second-best results across all four detectors. It ranks first on all three metrics for RT, TabFormer, and GraphSAGE, highlighting the value of its generated sequences for detectors that capture sequential or relational dependencies. 
On our production detector RT, PaySeqGen improves AUC by 0.393 percentage points over training without augmentation, exceeding the strongest generation baseline by 0.224 points. It also achieves the largest Recall@2 gain of 1.621 points, together with a KS increase of 0.218 points. Compared with real-fraud replay, PaySeqGen provides an additional 0.527 points of Recall@2, supporting the utility of generated samples beyond increasing the proportion of fraud examples in training.
Large transaction volumes in web payment systems make it important to prioritize high-risk payments for further assessment. The Recall@2 gain therefore indicates that PaySeqGen helps identify more fraud among these prioritized payments, helping platforms allocate limited risk-control resources more effectively.
The benefits are also evident in graph-based detection. On GraphSAGE, the Recall@2 gain reaches 4.561 points, exceeding the strongest generation baseline on this metric by 1.186 points. The accompanying AUC and KS improvements extend the evidence for useful augmentation to a detector that models relationships among payments.
On XGBoost, PaySeqGen ranks second to real-fraud replay on AUC
and KS, while outperforming all generation baselines, suggesting that its generated sequences also capture useful attribute-level patterns for fraud detection. 
Together, these findings support our fraud-policy reconstruction as a practical way to convert observed fraud behavior into reusable training resources for large-scale payment risk control, strengthening online payment security.

\begin{figure*}
    \centering
    \includegraphics[width=\textwidth]{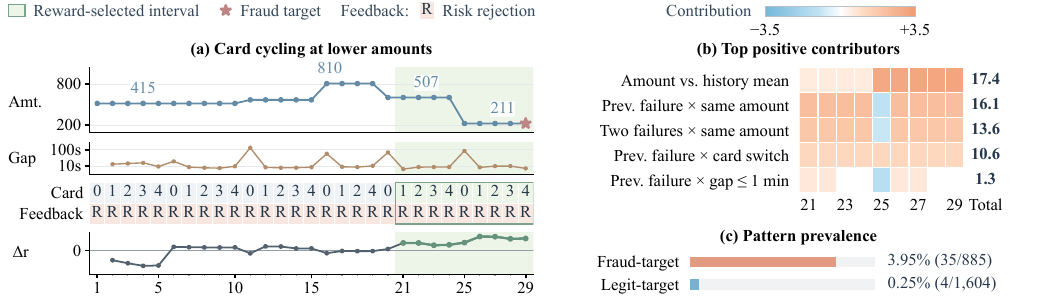}
    \vspace{-12pt}
    \caption{Reward-guided fraud-pattern analysis. (a) Payment history and relative rewards, with the selected interval shaded and the fraudulent target marked by a star. (b) Features with the largest positive contributions to relative reward within the selected interval. (c) Prevalence of the corresponding pattern among real fraud-target and legitimate-target trajectories.}
    \label{fig:case study}
\end{figure*}

\subsection{Ablation Study}

To assess the candidate construction mechanism and behavioral reward design, we compare three variants: \textit{w/o coh.} removes history conditioning from transition sampling while retaining joint payment changes; \textit{w/o expl.} replaces exploratory candidates with history-guided candidates while keeping the candidate count unchanged; and \textit{w/o inter.} removes feedback--action interactions from the reward while retaining all policy inputs. These comparisons assess how each component contributes to downstream detection performance.
All three ablations generally weaken performance relative to the full model, although their relative effects vary across detectors and metrics. Removing history conditioning reduces Recall@2 by 0.450 percentage points on RT and 3.135 points on GraphSAGE, suggesting that aligning payment proposals with their preceding histories is important for generating useful fraud examples. Removing exploration produces the largest or second-largest Recall@2 decline on every detector, including the largest decrease on RT (1.053 points). This is consistent with its role in broadening behavioral coverage for detector training. Removing interaction reward leads to the largest AUC decrease on three of the four detectors, suggesting that modeling contextual preferences is important for reconstructing behaviors that support accurate risk ranking.
Together, these findings support combining history-guided sampling, exploratory sampling, and feedback--action interaction reward to generate useful training sequences for payment fraud detection.

\subsection{Parameter Sensitivity}
\label{param sensitivity}
\begin{figure}
 \centering
 \includegraphics[width=\linewidth]{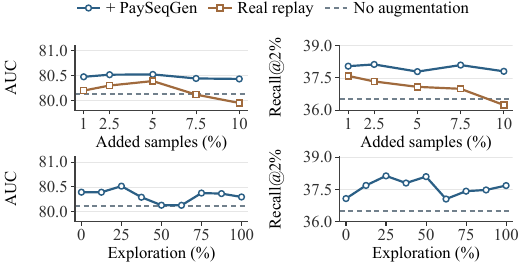}
 \vspace{-15pt}
 \caption{Sensitivity Analysis with respect to the ratio of augmented samples and the exploration candidate ratio.}
 \label{fig:sensitivity}
 \vspace{-15pt}
\end{figure}

We further investigate the effects of augmentation size and the proportion of exploratory candidates. As shown in Figure~\ref{fig:sensitivity}, PaySeqGen achieves higher AUC than real-fraud replay across the tested augmentation ratios, supporting the utility of generated samples beyond oversampling existing fraud examples. Its AUC increases as the augmentation ratio rises from 1\% to 5\%, whereas Recall@2 peaks at 2.5\%. Increasing the ratio beyond 5\% provides no consistent additional benefit. The 2.5\% setting thus combines the highest observed Recall@2 with near-peak AUC and a modest increase in training data, illustrating the value of a limited augmentation budget.
For candidate construction, a 25\% exploration ratio achieves the highest AUC and Recall@2 among the tested settings. Performance fluctuates at higher ratios, suggesting that increasing exploration alone does not consistently improve detector augmentation. This may reflect a trade-off between behavioral coverage and coherence: insufficient exploration may restrict the diversity of generated behaviors, while excessive exploration may introduce candidates that are less consistent with the observed payment history, adding noise to the generated sequences. These results support combining history-guided and exploratory candidates to generate useful training sequences for payment fraud detection.

\subsection{Interpretability Analysis}

We use a real payment trajectory with a fraudulent target to
illustrate reward-guided fraud-pattern analysis. In Figure~\ref{fig:case study}, consecutive positive relative rewards (Eq.~\eqref{eq:relative_reward}) highlight an interval in which the observed actions receive higher scores than the candidate mean under the same history. Within this interval, the payment records show rapid retries across multiple cards at a fixed amount, followed by further retries at a lower fixed amount. The heatmap presents top positive contributions to relative reward, including amounts relative to the historical mean and interactions between preceding failures and same-amount retries, card switching, or short payment intervals. Together, the payment records and reward contributions provide traceable evidence for interpreting how payment behavior evolves following repeated rejections.
To examine the pattern beyond this case, we further assess the prevalence of such sequence pattern using the reward-derived rule (specified in Appendix~\ref{app:pattern_prevalence}). We apply this rule to all eligible trajectories in the validation cohort without reward-based preselection. It matches 3.95\% of fraud-target trajectories and 0.25\% of legitimate-target trajectories. The higher prevalence in fraud-target trajectories suggests that the pattern is associated with fraudulent payment behavior. 
These findings demonstrate how PaySeqGen's interpretable behavioral cues can be used to help analysts focus their review of complex payment histories in large-scale online payment systems.

\subsection{Industrial Deployment and A/B Test}

\begin{figure}
    \centering
    \includegraphics[width=\linewidth]{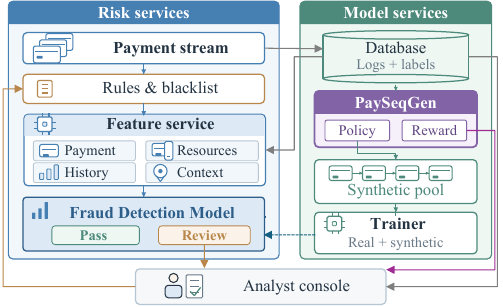}
    \vspace{-20pt}
    \caption{Schematic overview of online payment risk-control pipeline integrating risk services and model services.}
    \label{fig:deployment}
\end{figure}

PaySeqGen is integrated into Ant International's production online payment risk-control system to strengthen existing fraud detectors through data augmentation. As shown in Figure~\ref{fig:deployment}, risk services first perform rule and blacklist checks, followed by feature extraction and model-based risk scoring. Risk services log payment records and outcomes in the database. Model services use these records and fraud labels to reconstruct the policy and generate a synthetic sample pool. Sequences generated by PaySeqGen are combined with real payment records to retrain and update the detectors, which are deployed to risk services for subsequent payment screening. Behavioral cues derived from the learned reward are presented in the analyst console to assist analysts in fraud risk analysis.

Following deployment, we evaluate the resulting augmentation through online A/B tests on two production detectors within a segment of payment traffic over a 30-day period. We track both detection metrics and operational outcomes  throughout the evaluation period, including additional fraud identification and the associated reduction in losses. As shown in Table~\ref{tab:ab_test}, PaySeqGen augmentation improves all three metrics for both detectors. For Model-1 and Model-2, AUC increases by 0.21 and 0.26 percentage points, KS by 0.76 and 1.06 points, and Recall@2 by 3.19 and 2.84 points, respectively. The recall gains indicate that more fraudulent payments are captured at the same 2\% screening rate. Beyond these metric improvements, the results show that, compared with the system without augmentation, PaySeqGen augmentation helps identify 250 additional users involved in fraud and detect 393 additional fraudulent payments, preventing an additional US\$9,738 in losses. These results demonstrate the practical value of PaySeqGen for improving fraud detection performance and reducing financial losses in online payment services.

\begin{table}[t]
    \centering
    \vspace{-5pt}
    \caption{Results of online A/B tests.}
    \vspace{-10pt}
    \label{tab:ab_test}
    \setlength{\tabcolsep}{5pt}
    \begin{tabular}{l|c|c|c}
        \toprule
        Method & AUC & Recall@2 & KS \\
        \midrule
        Model-1 & 86.02 & 40.66 & 54.69 \\
        Model-1 (PaySeqGen) & 86.23 & 43.85 & 55.45 \\
        Model-2 & 85.90 & 41.79 & 55.03 \\
        Model-2 (PaySeqGen) & 86.16 & 44.63 & 56.09 \\
        \bottomrule
    \end{tabular}
    \vspace{-5pt}
\end{table}

\section{Related Work}

\subsection{Generative Data Augmentation}
Generative data augmentation operates at both the record and sequence levels to provide additional synthetic samples for model training.
Record-level generative models synthesize samples by learning dependencies among heterogeneous attributes~\cite{xu2019modeling,zhang2024mixedtype,shi2025tabdiff,lin2025ctsyn,shankar2026harpoon}, while fraud-oriented approaches use adversarial or diffusion models to augment minority-class transactions~\cite{11081459,MISHRA2026102716}. 
Beyond individual transactions, fraud patterns also unfold across successive payment attempts, where temporal dependencies and evolving choices carry information that isolated records cannot preserve. This motivates sequence-level generation that captures temporal structure~\cite{NEURIPS2019_c9efe5f2,yuan2024diffusion,NEURIPS2025_70fb27f3,shen2026latenttodata}, with extensions incorporating contextual conditions~\cite{pmlr-v235-narasimhan24a} or counterfactual augmentation of minority-class sequences~\cite{10.1145/3711896.3737049}. Banksformer further specializes autoregressive generation for financial transaction sequences~\cite{10.1007/978-3-031-26422-1_8}.
However, these approaches do not explicitly model the action--feedback loop through which payment outcomes guide subsequent decisions.
PaySeqGen instead reconstructs a policy from real demonstrations to generate payment sequences through this interaction process.

\vspace{-5pt}
\subsection{Policy Learning for Sequence Generation}

Modeling this interaction requires specifying or learning policies. AMLworld simulates financial activity through empirically calibrated agent mechanisms~\cite{NEURIPS2023_5f38404e}. To learn from observed behavior, behavior cloning and its extensions imitate expert choices and behavioral styles~\cite{osa2018algorithmic,yang2025diverse}. Beyond direct imitation, adversarial methods guide policy learning and sequence generation with discriminator rewards~\cite{NIPS2016_cc7e2b87,3298483.3298649}. For fraud generation, Fraud-RLA uses PPO with task-specific rewards and learns action covariance to model attribute dependencies~\cite{11434559}.
For payment fraud reconstruction, the behavioral reward is unknown, and predefined objectives may not reflect demonstrated preferences. IRL instead infers rewards that explain demonstrations and guide policy learning~\cite{1620270.1620297,pmlr-v48-finn16,garg2021iqlearn,pmlr-v267-ke25b,3780338.3782798,ICLR2026_194b8dac}. For example, MoE-BiEntIRL reconstructs graph-attack policies with interpretable rewards over graph and node features to generate adversarial samples~\cite{lyu2024enhancing}.
However, using IRL to reconstruct fraudulent payment strategies for sequence generation remains underexplored. This requires constructing coherent payment actions from interdependent attributes and learning behavioral preferences conditioned on preceding payment outcomes. PaySeqGen therefore constructs a finite candidate list through history-guided sampling of observed joint payment changes and exploratory sampling. Its decomposable behavioral reward models feedback--action interactions, allowing the same action to score differently after different outcomes. Together, these designs support interpretable reconstruction of feedback-driven fraud strategies, providing additional training sequences and traceable evidence for online payment risk analysis.

\vspace{-5pt}
\section{Conclusion}

In this paper, we propose PaySeqGen, an interpretable inverse reinforcement learning framework for reconstructing fraudulent payment strategies from observed online payment sequences. Specifically, our candidate construction mechanism combines history-guided and exploratory sampling to balance behavioral coherence and exploration. We further learn a decomposable linear behavioral reward that captures context-dependent preferences through feedback--action interactions. Through alternating reward learning and policy optimization, PaySeqGen translates observed behavior into additional training sequences and interpretable evidence for sequential fraud-pattern analysis. Experiments on large-scale real-world payment data demonstrate the utility of its generated sequences across multiple detector architectures, while a case study illustrates how reward contributions support the interpretation of multistep fraud behavior. Online A/B tests in Ant International's large-scale payment system further show Recall@2 gains of 3.19 and 2.84 percentage points for two production fraud detectors. These findings highlight PaySeqGen's practical value for strengthening fraud detection and supporting the analysis of complex fraud patterns in real-world web payment services.





\normalem
\bibliographystyle{ACM-Reference-Format}
\bibliography{references}

\appendix
\section{Pseudocode of PaySeqGen}
Algorithm~\ref{alg:payseqgen} summarizes alternating reward and policy learning. Expert and generated continuations share the same observed prefix. We write $\bar f(\tau)=L^{-1}\sum_{t=p+1}^{T}f(s_t,a_t)$ for the average continuation features, where $L=T-p$. Rollouts are applied to each prefix in the sampled minibatch, and $\eta_\theta$ denotes the reward learning rate. The payment feedback interface and candidate sampling rules remain fixed throughout training.

\begin{algorithm}[t]
\caption{PaySeqGen training}
\label{alg:payseqgen}
\begin{algorithmic}[1]
\Statex \textbf{Input:} Demonstrations $\mathcal D$; fixed feedback interface $P_{\mathrm{env}}$; candidate counts $K_h,K_e$; policy updates per reward update $U$
\Statex \textbf{Output:} Policy $\pi_\omega$ and reward $r_\theta$
\State Precompute joint changes between consecutive payments in $\mathcal D$
\State Initialize Actor--Critic parameters $\omega,\psi$; set $\theta\gets\mathbf 0$
\State Initialize $\pi_\omega$ by behavior cloning with demonstrated actions included in candidate lists
\For{each training round}
    \State Sample expert trajectories $\mathcal E$ and initial prefixes
    \State Replay expert continuations with recorded feedback
    \State $\mathcal P\gets\Call{Rollout}{\pi_\omega,\text{prefixes}}$
    \State $\mu^E\gets |\mathcal E|^{-1}\sum_{\tau\in\mathcal E}\bar f(\tau)$
    \State $\mu^\pi\gets |\mathcal P|^{-1}\sum_{\tau\in\mathcal P}\bar f(\tau)$
    \State $g_\theta\gets\mu^E-\mu^\pi-\lambda\theta$
    \State $\theta\gets\operatorname{clip}(\theta+\eta_\theta g_\theta,-w_{\max},w_{\max})$
    \For{$u=1$ \textbf{to} $U$}
        \State Sample fresh $\mathcal P\gets\Call{Rollout}{\pi_\omega,\text{prefixes}}$
        \State Compute $\tilde r_t$ using Eq.~\eqref{eq:training_reward}, with $\theta$ fixed
        \State Compute $\widehat A_t,\widehat G_t$ using Eq.~\eqref{eq:appendix_gae}
        \State Update $\omega,\psi$ using Eq.~\eqref{eq:rl_training_loss}, with $\widehat A_t,\widehat G_t$ detached
    \EndFor
\EndFor
\State \Return $\pi_\omega$ and $r_\theta(s,a)=\theta^\top f(s,a)$
\Statex
\Function{Rollout}{$\pi_\omega,\text{prefixes}$}
    \For{each remaining payment step}
        \State Construct $K_h$ joint history-guided actions using Eq.~\eqref{eq:historical_action_sampling}
        \State Add $K_e$ constrained exploration actions to form $\mathcal A_t$
        \State Sample $j_t\sim\pi_\omega(\cdot\mid s_t,\mathcal A_t)$; set $a_t\gets a_{t,j_t}$
        \State Sample $o_t\sim P_{\mathrm{env}}(\cdot\mid s_t,a_t)$
        \State Record the decision; $s_{t+1}\gets\operatorname{Update}(s_t,a_t,o_t)$
    \EndFor
    \State \Return generated continuations and decision records
\EndFunction
\end{algorithmic}
\end{algorithm}

\paragraph{Advantage and return estimation.}
For a completed rollout, abbreviate the Critic prediction as $V_t=V_\psi(s_t,\mathcal A_t)$. With terminal values $V_{T+1}=\widehat A_{T+1}=0$, GAE computes the following quantities backward:
\begin{equation}
\label{eq:appendix_gae}
\begin{aligned}
e_t &= \tilde r_t+\gamma V_{t+1}-V_t,\\
\widehat A_t &= e_t+\gamma\lambda_{\mathrm{GAE}}\widehat A_{t+1},\\
\widehat G_t &= \widehat A_t+V_t.
\end{aligned}
\end{equation}
Here, $e_t$ is the temporal-difference residual. The discount factor $\gamma$ controls the weight of future rewards, while $\lambda_{\mathrm{GAE}}$ controls how strongly later residuals contribute to the advantage. Both targets are detached during optimization. Entropy is already included in $\tilde r_t$ and is not added again to the loss.

\section{Details of Baselines}

We compare PaySeqGen with five generation baselines and a real-fraud replay control, and evaluate their augmentation effects on four target detectors.

\subsection{Generation Baselines}

The distribution-based baselines model the joint structure of payment records or sequences:
\begin{itemize}
    \item \textbf{CTGAN}~\cite{xu2019modeling} uses conditional adversarial learning to capture dependencies among continuous and categorical attributes. We adapt it to generate payment continuations conditioned on an observed prefix.
    \item \textbf{Diffusion-TS}~\cite{yuan2024diffusion} generates multivariate time series through diffusion, combining temporal decomposition with reconstruction-based learning to capture sequence structure. We use conditional infilling to generate payment continuations.
    \item \textbf{TabDiff}~\cite{shi2025tabdiff} models numerical and categorical attributes through a unified diffusion framework with learnable noise schedules. We apply it to mixed-type payment sequences while keeping observed prefix fields fixed.
\end{itemize}

The policy-based baselines learn how to select successive payment actions:
\begin{itemize}
    \item \textbf{BC}~\cite{osa2018algorithmic} learns a policy through supervised imitation of demonstrated actions. It uses PaySeqGen's policy architecture and generation candidate mechanism, but is trained independently without IRL reward learning or subsequent RL optimization.
    \item \textbf{MoE-BiEntIRL}~\cite{lyu2024enhancing} combines mixture-of-experts modeling with IRL to reconstruct attack policies and explain behavioral preferences. We adapt its graph-specific action representation to complete payment actions.
\end{itemize}

We additionally include \textbf{Real-fraud replay}, which appends an equal number of resampled real fraud-target examples. This control distinguishes the benefits of generating behavioral variations from increasing the training weight of observed fraud examples.

\begin{table*}[t]
\centering
\caption{Augmentation results on public credit card fraud detection dataset MS-FFSD.}
\vspace{-10pt}
\label{tab:msffsd_results}
\setlength{\tabcolsep}{2.5pt}
\begin{tabular}{@{}l|rrr|rrr|rrr|rrr@{}}
\toprule
Augmentation & \multicolumn{3}{c|}{RT} & \multicolumn{3}{c|}{TabFormer} & \multicolumn{3}{c|}{GraphSAGE} & \multicolumn{3}{c}{XGBoost} \\
\midrule
& $\Delta$AUC & $\Delta$Recall@2 & $\Delta$KS
& $\Delta$AUC & $\Delta$Recall@2 & $\Delta$KS
& $\Delta$AUC & $\Delta$Recall@2 & $\Delta$KS
& $\Delta$AUC & $\Delta$Recall@2 & $\Delta$KS \\
\midrule
Real-fraud replay &  -3.874 & \underline{-0.654} & \underline{-2.781} & +0.249 & -0.292 & +0.159 & -1.221 & \textbf{-0.905} & -0.327 & -0.874 & -0.028 & -0.242 \\
\midrule
BC & -11.970 & -0.710 & -10.376 & -1.583 & -0.306 & -1.531 & -0.439 & -1.184 & +0.708 & -1.175 & +0.084 & -1.019 \\
CTGAN & -12.406 & -1.058 & -11.004 & -4.368 & -0.794 & -3.842 & -6.084 & -1.351 & -5.278 & \underline{-0.404} & \textbf{+0.292} & -0.234 \\
Diffusion-TS & -4.191 & -0.947 & -2.809 & \underline{+0.449} & \underline{-0.111} & \underline{+0.559} & -0.548 & -1.253 & +0.258 & -0.894 & +0.014 & -0.401 \\
TabDiff & \textbf{-2.361} & -0.961 & -3.202 & -4.600 & -0.473 & -4.566 & -3.620 & -1.392 & -3.241 & \textbf{-0.337} & \underline{+0.265} & \underline{-0.150} \\
MoE-BiEntIRL & -9.098 & \underline{-0.654} & -8.137 & +0.157 & \textbf{-0.097} & +0.201 & \underline{-0.399} & \underline{-1.003} & \underline{+0.960} & -0.609 & 0.000 & \textbf{-0.119} \\
\midrule
PaySeqGen & \underline{-2.528} & \textbf{-0.432} & \textbf{-1.748} & \textbf{+1.011} & -0.223 & \textbf{+1.132} & \textbf{+2.194} & -1.267 & \textbf{+2.825} & -0.704 & -0.097 & -0.486 \\
\bottomrule
\end{tabular}
\end{table*}

\subsection{Target Models}

We use four detectors to assess augmentation across sequence, graph, and tree architectures:
\begin{itemize}
    \item \textbf{RT}, our production Transformer-based payment fraud detector, serves as the primary target model. It models historical payment sequences to assess fraud risk.
    \item \textbf{TabFormer}~\cite{padhi2021tabular} models payment sequences using hierarchical Transformers at the field and payment-event levels. It performs masked-attribute pretraining followed by fine-tuning for fraud detection.
    \item \textbf{GraphSAGE}~\cite{hamilton2017inductive} learns node representations by aggregating neighborhood information. We construct temporal, same-card, and same-device relations within payment histories to capture relationships among payments.
    \item \textbf{XGBoost}~\cite{chen2016xgboost} is a well-established tree-based baseline in production fraud detection. It predicts risk from summary features of the payment history.
\end{itemize}

\section{Results on Public Dataset}
\label{app:public_data}

We additionally evaluate on MS-FFSD~\cite{ref38,ref40}, a public financial transaction dataset with semantic enrichment and synthesized timestamps.
The released data contain 77,881 payment records from 30,346 users. We split events chronologically in proportions of approximately 60\%/20\%/20\%, keeping equal timestamps in the same split. Only explicitly legitimate or fraudulent targets contribute to detector supervision; unlabeled events may appear in the preceding history. This yields 16,965 training, 6,519 validation, and 6,159 test targets. Detectors use at most 19 strictly preceding payments, without target attributes or historical fraud labels, keeping the same with main experiment.
Generator training is restricted to users with more than 20 training-period payments, yielding 3,662 complete windows, including 109 fraud-target windows. Each method adds 2,090 fraud-target examples, approximately 12\% of the original detector training set. 
MS-FFSD provides merchant, location, transaction type, province, amount, and time, but lacks payment outcomes and reliable card/device identifiers. We therefore omit the payment feedback interface and feedback--action interactions and use 34 fixed features describing amounts, intervals, and merchant/location usage. The adapted policy still selects from 48 history-guided and 16 exploration candidates. 

Table~\ref{tab:msffsd_results} reports changes relative to the unaugmented detectors. Many augmentation methods yield negative gains on this dataset. For generators, this may be partly due to the limited availability of long payment histories and the small number of fraud-target windows available for learning sequential behavioral patterns. Despite these constraints and the absence of feedback-related components, PaySeqGen remains competitive, achieving the best or second-best AUC and KS on RT, TabFormer, and GraphSAGE. These results support PaySeqGen's adaptability to financial datasets with limited behavioral information, even without payment feedback or feedback--action interactions.

\section{Complexity analysis}
\label{app:complexity}

Let $n$ be the maximum visible history length, $K$ the candidate count, and $N$ the number of training payment transitions. With fixed feature dimensions, network sizes, relation types, and path depth, history processing and policy scoring cost $O(n^2+Kn)$. After preprocessing, indexed retrieval and candidate adaptation cost $C_{\mathrm{cand}}=O(K_h\log N+Kn)$, with fixed fallback and resampling limits. The feedback interface caches its Transformer history representation and scores candidates through a shared prediction head, giving $C_{\mathrm{env}}=O(n^2+K)$. The total per-decision cost is therefore
\begin{equation}
\label{eq:appendix_complexity}
C_{\mathrm{step}}
=
O(n^2+Kn)+C_{\mathrm{cand}}+C_{\mathrm{env}}.
\end{equation}
For $S$ decision steps processed during training, including expert replay and policy rollouts, the total cost is $O(SC_{\mathrm{step}})$ after preprocessing and initialization. Generating an $L$-step continuation costs $O(LC_{\mathrm{step}})$. Reward updates require only linear aggregation of precomputed features, avoiding the expert-responsibility and gating updates required by MoE-BiEntIRL~\cite{lyu2024enhancing}. Thus, reward learning preserves the asymptotic order of the underlying Actor--Critic computation.
Training and generation take place during detector preparation and add no generation overhead to real-time payment scoring, supporting practical integration into online payment risk-control systems.

\section{Details of Fraud-Pattern Prevalence}
\label{app:pattern_prevalence}

\paragraph{Reward-derived rule.}
We derive a behavioral rule from the interval highlighted by PaySeqGen's relative rewards in  Figure~\ref{fig:case study} and the corresponding feature contributions. The contributions link same-amount retries and card switching to preceding failures, while the payment records reveal two consecutive stages with lower amounts in the second stage. We encode this evidence as a deterministic window-matching rule.
For a payment interval $I$, let $\ell_i=\log(1+m_i)$ and $N_{\mathrm{sw}}(I)$ count changes between consecutive cards. We define the stage-validity predicate:
\begin{equation}
\label{eq:pattern_stage}
\begin{aligned}
\operatorname{Valid}(I)
={}&[N_{\mathrm{sw}}(I)\ge 2] 
\land
\left[
\max_{i\in I}\ell_i-\min_{i\in I}\ell_i
\le \epsilon
\right],
\end{aligned}
\end{equation}
where $\epsilon=10^{-4}$.
We scan windows of 6--12 consecutive payments, excluding windows with missing card identifiers. Each window is partitioned into two consecutive stages containing at least three payments each. For a window $W$ of $n$ locally indexed payments, each split $k=3,\ldots,n-3$ defines $I_1=\{1,\ldots,k\}$ and $I_2=\{k+1,\ldots,n\}$.
The matching predicate is:
\begin{equation}
\label{eq:pattern_match}
\begin{aligned}
\operatorname{Match}_{\kappa}(W,k)
={}&\operatorname{Valid}(I_1)
\land\operatorname{Valid}(I_2)  \land\left[\min_{i\in I_1}m_i>0\right] \\
\land&\left[
\max_{i\in I_2}m_i
\le (1-\kappa)\min_{i\in I_1}m_i
\right] \land
\bigwedge_{i=1}^{n-1}
[o_i\in\mathcal O_{\mathrm{fail}}],
\end{aligned}
\end{equation}
where $\mathcal O_{\mathrm{fail}}$ contains failure outcomes
and $\kappa=0.2$ specifies the minimum amount reduction.
A trajectory matches if this predicate holds for any window
and split. The final outcome within each window is unused,
and no additional constraints are imposed on device identity,
failure subtype, or payment intervals.

\paragraph{Prevalence analysis.}
The matching criteria and thresholds were fixed before
cohort-wide counting.
Table~\ref{tab:pattern_prevalence_sensitivity} reports prevalence
under minimum amount reductions of 10\%, 20\%, and 30\%.
The pattern is consistently more prevalent among fraud-target
trajectories, suggesting that the observed association is not
specific to the default 20\% threshold.
These results provide exploratory evidence that the
reward-highlighted pattern recurs beyond the illustrated case
and is associated with fraudulent targets.

\begin{table}
\centering
\caption{Pattern prevalence under different minimum amount reductions. Entries show matched/eligible trajectories and occurrence rates.}
\vspace{-10pt}
\label{tab:pattern_prevalence_sensitivity}
\setlength{\tabcolsep}{5pt}
\begin{tabular}{lrr}
\toprule
Reduction & Fraud-target & Legitimate-target \\
\midrule
10\% & 43/885 (4.86\%) & 5/1,604 (0.31\%) \\
20\% & 35/885 (3.95\%) & 4/1,604 (0.25\%) \\
30\% & 31/885 (3.50\%) & 3/1,604 (0.19\%) \\
\bottomrule
\end{tabular}
\end{table}

\section{Quality of Generated Fraud Samples}
\label{app:fraud_sample_quality}

We compare PaySeqGen with generation baselines on behavioral fidelity and coverage. We derive fixed trajectory descriptors from payment attributes and outcomes, summarizing local behavior, feedback-dependent transitions, and card/device reuse structure.

\begin{itemize}
    \item \textbf{Local W1.}
    We average the 1-Wasserstein distances~\cite{peyre2019computational} between real and generated distributions of trajectory-level statistics, with each distance normalized using its training-data scale.
    
    \item \textbf{Conditional and two-step JS.}
    Conditional Jensen--Shannon (JS) divergence~\cite{lin1991divergence} compares distributions of discretized action categories conditioned on preceding feedback, weighted by real feedback frequencies. Two-step JS compares consecutive feedback--resource-change patterns separately for cards and devices.
    
    \item \textbf{Path MMD$^2$.}
    Within each trajectory, directed edges connect payments to later payments sharing a card or device. We compare distributions of normalized path counts for paths with one to three edges using squared maximum mean discrepancy with a Gaussian kernel~\cite{gretton2012kernel}.
    
    \item \textbf{Coverage.}
    Following Naeem et al.~\cite{naeem2020reliable}, coverage is the fraction of real trajectory representations whose neighborhoods contain at least one generated representation. Representations combine the descriptors above. Neighborhood radii are distances to the $k$th nearest other real representation, with $k=5$.
\end{itemize}

Lower discrepancy scores indicate closer agreement with real behavioral statistics, while higher coverage indicates that more real trajectories have nearby generated samples in the descriptor space.

\begin{table}[t]
\centering
\small
\caption{Behavioral fidelity and coverage of generated fraud samples on the proprietary dataset. Values are means; bold denotes the best value in each column.}
\vspace{-10pt}
\label{tab:fraud_sample_quality}
\setlength{\tabcolsep}{3pt}
\begin{tabular}{lrrrrr}
\toprule
Method & \shortstack{Local\\W1 $\downarrow$} & \shortstack{Cond.\\JS $\downarrow$} & \shortstack{2-step\\JS $\downarrow$} & \shortstack{Path\\MMD$^2$ $\downarrow$} & \shortstack{Cov.\\(\%) $\uparrow$} \\
\midrule
BC & \textbf{0.316} & \textbf{0.180} & 0.486  & \textbf{0.0313} & \textbf{24.44} \\
CTGAN & 1.462 & 0.380 & 0.349 & 0.2352 & 1.39 \\
Diffusion-TS & 0.613 & \uline{0.299} & 0.284 & 0.0574 & 10.21 \\
TabDiff & 1.086 & 0.335 & \uline{0.240} & \uline{0.0373} & 0.98 \\
MoE-BiEntIRL & 0.730 & 0.451 & 0.472 & 0.1407 & 4.97 \\
\midrule
PaySeqGen  & \uline{0.560} & 0.369 & \textbf{0.176} & 0.0540 & \uline{18.12} \\
\bottomrule
\end{tabular}
\end{table}

Table~\ref{tab:fraud_sample_quality} shows that PaySeqGen
achieves the lowest two-step JS, second-lowest Local W1,
and second-highest coverage. Two-step JS evaluates consecutive
feedback--resource-change patterns, making it particularly
relevant to reconstructing sequential payment behavior.
PaySeqGen's advantage on this metric suggests that it better
preserves these short behavioral patterns, providing evidence
for its effectiveness in modeling feedback-dependent decisions.
Performance nevertheless varies across metrics, with BC
achieving the best results on the other four measures.
These metrics assess agreement with selected reference
statistics and neighborhood coverage in the chosen descriptor
space. Such agreement alone does not establish whether
generated samples introduce useful behavioral variations
for detector training. We therefore use downstream detection
performance as the primary evaluation criterion and report
these metrics as complementary diagnostics. Together, the
results provide supporting evidence of generated-sample
fidelity, particularly in consecutive behavioral patterns
relevant to fraud-strategy reconstruction.

\end{document}